\documentclass[letterpaper]{article} 
\usepackage{aaai2027}
\usepackage[hyphens]{url}  
\usepackage{graphicx} 
\usepackage{amsfonts}
\usepackage{amssymb} 
\usepackage[numbers,sort&compress]{natbib}

\usepackage{caption} 
\usepackage{algorithm}
\usepackage{algorithmic}

\usepackage{pifont}
\usepackage{multirow}
\usepackage{amsmath}
\usepackage{newfloat}
\usepackage{listings}
\usepackage{cuted}
\DeclareCaptionStyle{ruled}{labelfont=normalfont,labelsep=colon,strut=off} 
\floatstyle{ruled}
\newfloat{listing}{tb}{lst}{}
\floatname{listing}{Listing}

\usepackage{booktabs}

\title{Is It Time for the Renaissance of Salient Object Detection in the Era of MLLMs?}
\author{
    Wenzhuo Zhao\textsuperscript{\rm 1},
    Xiuzhi Li\textsuperscript{\rm 1},
    Zhongkuan Mao\textsuperscript{\rm 1},
    Ronghao Xian\textsuperscript{\rm 1},
    Yao Jiang\textsuperscript{\rm 1},\\
    Zhao Gao\textsuperscript{\rm 1},
    Keren Fu\textsuperscript{\rm 1,\rm 2}\corresponding,
    Qijun Zhao\textsuperscript{\rm 1,\rm 2},
    Jian Cheng\textsuperscript{\rm 3}
}
\affiliations{
    \textsuperscript{\rm 1}College of Computer Science, Sichuan University, China\\
    \textsuperscript{\rm 2}National Key Laboratory of Fundamental Science on Synthetic Vision, Sichuan University, China\\
    \textsuperscript{\rm 3}School of Information and Communication Engineering, University of ElectronicScience and Technology of China, China\\
}

\begin{document}
\maketitle

\begingroup
\setlength{\stripsep}{0pt}

\begin{strip}
    \vspace*{-8mm}
    \centering

    \includegraphics[
        width=\textwidth
    ]{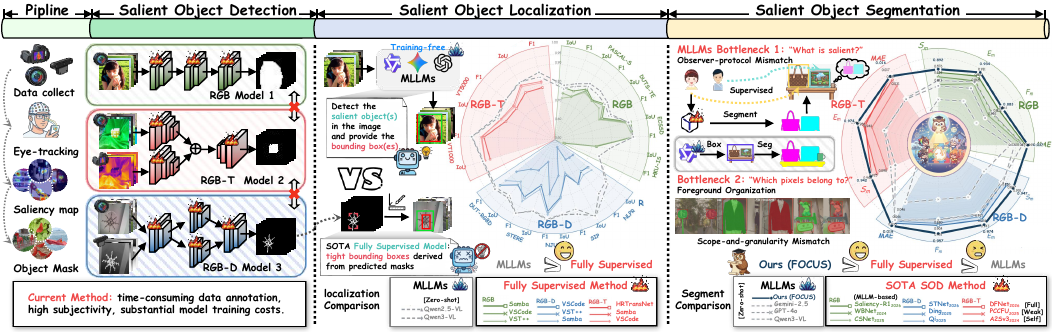}
    \captionsetup{
        aboveskip=6pt,
        belowskip=0pt
    }
    \vspace{-4mm}
    \captionof{figure}{Rethinking SOD with MLLMs: In contrast to annotation- and training-intensive SOD methods, our diagnostic benchmark (SaliLLM) reveals the superior salient-object perception capabilities of MLLMs, while our method addresses two key bottlenecks to outperform both fully supervised SOTA models and off-the-shelf MLLMs without task-specific training.}
    \label{fig_intro}
    \vspace{2mm}
\end{strip}
\endgroup

\begin{abstract}
The zero-shot capabilities of multimodal large language models (MLLMs) are pushing salient object detection (SOD) beyond task-specific supervision. To disentangle MLLMs beyond conventional mask-based evaluation, we decompose SOD into localization and segmentation, and re-engineer datasets with phrases, boxes, and attributes, establishing a diagnostic benchmark for MLLM saliency perception (SaliLLM).
SaliLLM uncovers a striking capability mismatch: MLLMs outperform state-of-the-art (SOTA) methods in localization, yet remain substantially weaker in segmentation. Further analyses attribute this gap primarily to mismatches between MLLMs and annotations over foreground cardinality, granularity, and extent.
Motivated by this diagnosis, we recast zero-shot SOD as protocol-aligned \textbf{F}oreground \textbf{O}rganization and introduce the first training-free framework that leverages Gestalt-inspired \textbf{C}ollaborative attention for \textbf{U}nified \textbf{S}OD (FOCUS).
FOCUS couples top-down Bayesian-surprise calibration of protocol-conditioned foreground granularity with bottom-up propagation of MLLMs evidence over entity-centric perceptual manifolds induced by self-supervised features, yielding coherent object extents as prompts for a general segmenter.
Across 13 RGB, RGB-D, and RGB-T SOD benchmarks, FOCUS generally surpasses SOTA methods without training, reducing mean absolute error by \textbf{11\%}, \textbf{34\%}, and \textbf{48\%} compared with fully, weakly, and self-supervised methods, respectively.
Our findings signal the renaissance of SOD: from task-specific supervision to zero-shot foreground organization. Code is available in the supplementary material.
\end{abstract}


\section{Introduction}
\label{sec:Introduction}
Salient Object Detection (SOD) aims to identify the entities that most attract an observer's attention and predict their pixel-level masks \cite{zhou2021rgb,luo2025vscode}. For decades, RGB, RGB-D, and RGB-T SOD has predominantly relied on manually constructed dense mask supervision and modality-specific architectures, as illustrated in Fig.~\ref{fig_intro} \cite{fan2019shifting,cong2025breaking,wan2026rsonet}. Although this paradigm has steadily improved detection accuracy, its progress remains fundamentally dependent on costly annotations, dataset-specific distributions, and task-specialized models. Meanwhile, Multimodal Large Language Models (MLLMs) have demonstrated remarkable zero-shot capabilities in visual understanding and object localization \cite{yang2026discover}. This raises a fundamental question: \emph{If foundation models can already identify salient entities in an image, does SOD still require task-specific supervision?}

Existing mask-level evaluation cannot answer this question, because the final mask conflates saliency reasoning, spatial localization, and pixel-level segmentation. Existing MLLM-based segmentation systems typically rely on additional post-training, while directly prompting an MLLM to generate masks likewise obscures its intrinsic saliency perception capability \cite{li2025saliency,tang2026ufo}.

To this end, we introduce \textbf{SaliLLM}, a diagnostic benchmark for disentangling saliency perception in MLLMs. SaliLLM decomposes SOD into \emph{salient entity localization} and \emph{prompted segmentation}, and augments existing multimodal SOD datasets with entity level phrases, tight bounding boxes, semantic categories, and saliency attributes. The benchmark reveals a striking capability mismatch. As shown in Fig.~\ref{fig_intro}, MLLMs already outperform state-of-the-art (SOTA) SOD methods, including Samba \cite{he2025samba}, VSCode \cite{luo2024vscode}, and VST++ \cite{liu2024vst++}, in box level salient entity localization, yet remain substantially weaker in segmentation. In other words, MLLMs already know \emph{where} the salient entities are, but still struggle to translate this knowledge into accurate foreground masks.

Layer-wise readouts and controlled interventions trace this gap primarily to mismatches between model predictions and annotations in foreground cardinality, granularity, and extent. (1) \textbf{Bottleneck--I: ``What is salient?'' Observer Protocol Mismatch.} Saliency is not an entirely objective property. Different datasets encode distinct preferences regarding foreground entities, granularity, and spatial extent. We recover this semantic prior only from original dataset papers and public acquisition or annotation descriptions; benchmark images and derived label statistics are excluded from protocol construction.
(2) \textbf{Bottleneck--II: ``Which pixels belong to it?'' Foreground Organization Mismatch.} Even when the correct entities are selected, the sparse semantic evidence provided by MLLMs does not naturally form complete and precise object extents. Their predictions typically exhibit \emph{high coverage but low purity}. Downstream promptable segmenters can refine local shapes and boundaries, but cannot rectify upstream errors in object granularity and extent.

This process echoes the Gestalt theory of visual perception, in which coherent objects emerge from the interplay between top down cognitive expectations and bottom-up perceptual grouping. Building on this insight, we recast zero--shot SOD as protocol-conditioned \textbf{F}oreground \textbf{O}rganization and introduce the first training-free framework that leverages Gestalt-inspired \textbf{C}ollaborative attention for \textbf{U}nified zero-shot \textbf{S}OD (FOCUS). 
Its top down observer protocol pathway uses Bayesian surprise to calibrate foreground granularity and spatial extent according to the observer protocols encoded by different datasets. Observer protocols are derived exclusively from public dataset papers and acquisition descriptions. In parallel, its bottom-up visual perception pathway propagates sparse semantic evidence from MLLMs over entity centered perceptual manifolds constructed from self supervised features. The two pathways jointly consolidate sparse attention into aligned object regions and semantic phrases, which are subsequently fed into a segmentation foundation model.

Across 13 RGB, RGB-D, and RGB-T benchmarks, FOCUS achieves the strongest overall performance among the compared training-free systems and is competitive with or superior to recent supervised SOTA models. These results provide the \textbf{first evidence} that visual knowledge in foundation models, once organized through protocol conditioned foreground organization, can surpass task-specific supervision. FOCUS marks a renaissance in SOD, shifting the field from learning dataset specific segmentation masks to organizing general foreground knowledge already encoded in foundation models. This result shows that protocol-conditioned organization can convert frozen foundation-model knowledge into a strong alternative to task-specific SOD training.

Our contributions are summarized as follows:
\begin{itemize}
\item We introduce the first diagnostic benchmark that disentangles salient entity localization from prompted segmentation for evaluating perception in MLLMs. It reveals a striking capability mismatch: MLLMs exhibit strong localization but remain substantially weak in segmentation.

\item We diagnose this gap through the internal saliency pathways of MLLMs and identify mismatches in foreground granularity, extent, and coherence as the bottleneck. This finding motivates a new formulation of zero-shot SOD as protocol conditioned foreground organization.

\item We propose the first training-free framework that unifies multi-modal SOD through top down protocol calibration and bottom-up perceptual propagation. FOCUS generally surpasses SOTA supervised methods across 13 datasets.

\end{itemize}

\section{Related Work}
\label{sec:Related Work}

\paragraph{Salient Object Detection.}
Conventional SOD relies on modality-specific architectures for dense prediction \cite{wang2021salient,zhou2021rgb}. Even weakly- and self-supervised methods require domain-specific training to alleviate pixel-level annotation costs. Recently, Saliency-R1 \cite{li2025saliency} demonstrated that fine-tuning MLLMs via reinforcement learning can instill saliency reasoning. In contrast, we investigate the intrinsic, zero-shot capabilities of off-the-shelf MLLMs—devoid of any SOD-specific training. Through a stage-wise evaluation, we isolate the critical gap between semantic saliency perception and dense prediction.

\paragraph{MLLMs and Promptable Segmentation.}
SAM \cite{kirillov2023segment} establishes a prompt-driven segmentation foundation, subsequently extended by Grounded SAM with open-vocabulary localization. Xia et al. further aligns language reasoning with pixel decoders for query-conditioned segmentation \cite{xia2024gsva}. Crucially, these interactive paradigms rely on explicit queries to specify the target, focusing entirely on \textit{how} to segment. SOD fundamentally differs: it requires autonomously determining \textit{what} is salient under observer protocols. FOCUS therefore decouples this pipeline. It concentrates primary modeling capacity on semantic foreground organization, relegating the promptable segmenter strictly to a pixel-level execution engine for protocol-aligned prompts.

\section{Benchmark and Analysis}
\label{sec:Benchmark_and_Analysis}

\begin{figure*}[!t]
    \centering
    \captionsetup{skip=5pt}
    \includegraphics[width=1\linewidth]{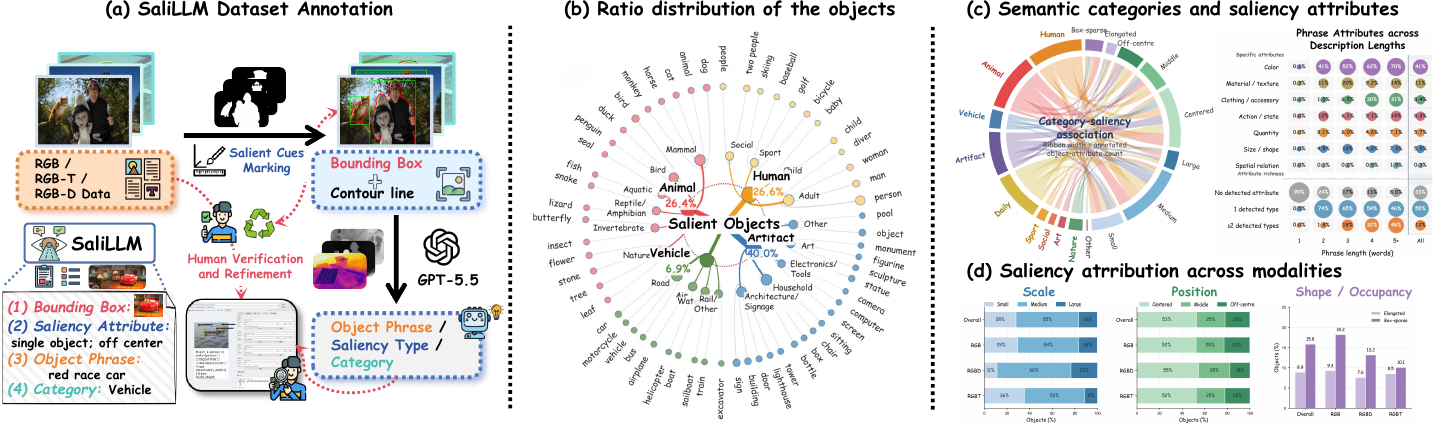}
    \caption{Overview of the SaliLLM benchmark: construction pipeline, semantic taxonomy, and dataset statistics.}
    \label{fig_bench}
\end{figure*}

\begin{figure*}[!t]
    \centering
    \captionsetup{skip=5pt}
    \includegraphics[width=1\linewidth]{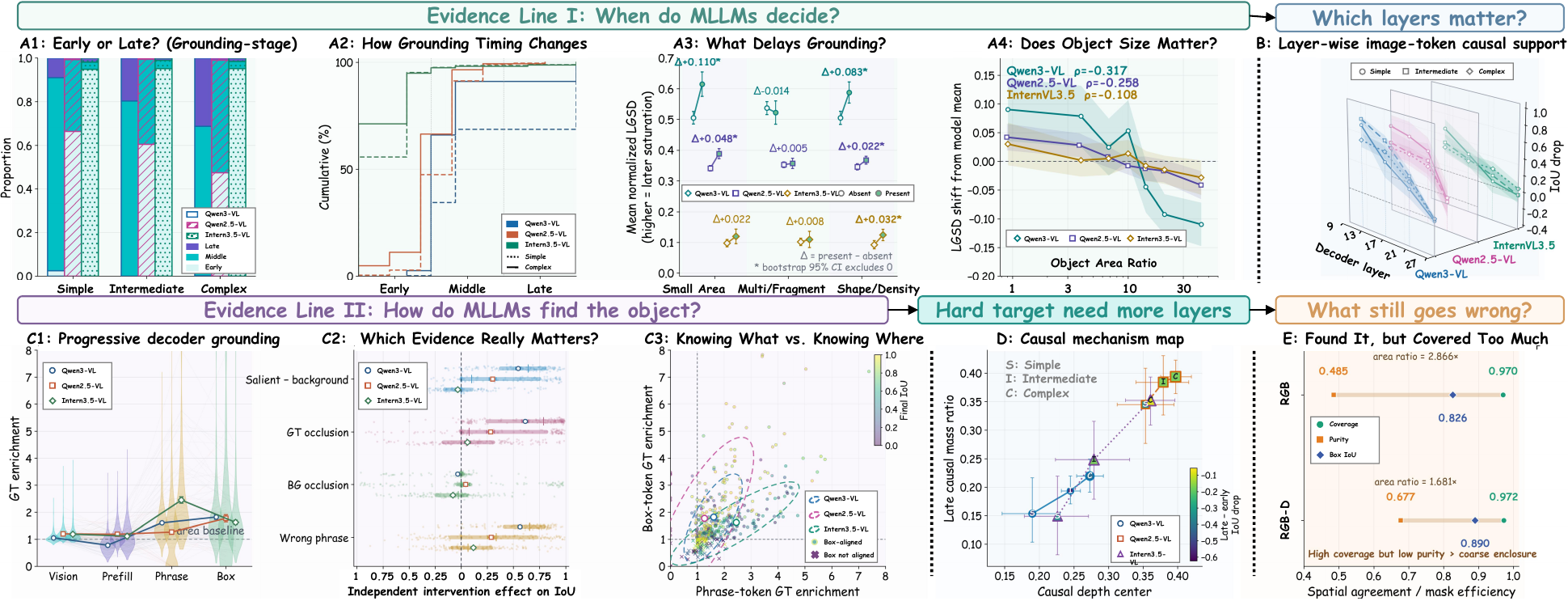}
    \caption{Analysis of MLLMs on SaliLLM: progressive salient object perception and incomplete spatial closure in segmentation.}
    \label{fig_analysis}
\end{figure*}

\subsection{SaliLLM Benchmark}
\label{sec:benchmark}
\subsubsection{Data Annotation}
SaliLLM is constructed from existing RGB, RGB-D, and RGB-T SOD datasets (see the appendix for details) and comprises 20,932 images and 32,040 salient entities, covering diverse imaging modalities, scene types, and object appearances. As illustrated in Fig.~\ref{fig_bench} (a), we adopt an iterative human-in-the-loop annotation pipeline that combines GPT-5.5 with human verification and refinement. A team of four human reviewers verified the annotations of all samples, including object phrases, bounding boxes, category labels, and saliency attributes. Disagreements were resolved through discussion and consensus.
For each salient entity, SaliLLM provides a semantic phrase, a bounding box, a category label, a set of saliency-related attributes, and a mask. We redistribute only derived annotations, leaving images and masks under their original licenses. These source masks are used to construct and evaluate SaliLLM annotations, but are never exposed to protocol construction or inference.

\subsubsection{Category and Saliency Attributes}
As shown in Fig.~\ref{fig_bench} (b), we calibrate the semantic annotations using the object phrases and organize all entities into ten unified categories. Each entity is further annotated with a set of saliency attributes, including \textit{small}, \textit{large}, \textit{centered}, \textit{off-center} and \textit{elongated}. These attributes characterize the entity in terms of scale and spatial distribution, enabling fine-grained analysis of model behavior across different saliency conditions.

\subsubsection{Dataset Features and Statistics}
As shown in Fig.~\ref{fig_bench} (c), categories, representative phrases, and saliency attributes exhibit rich hierarchical and correlation patterns. SaliLLM contains 9,812 unique object phrases, averaging 2.6 words in length. These phrases span people, animals, vehicles, natural objects, and diverse activities, while encoding discriminative cues such as color, action, and material to provide explicit semantic guidance for salient entity perception.

\begin{table}[t]
  \centering
  \small
    \caption{Phrase-level salient-entity perception and semantic--localization error decomposition.}
  \label{tab:phrase_recognition}
  \vspace{-1mm}
  \renewcommand{\arraystretch}{0.75}
  \setlength\tabcolsep{2.4pt}
  \begin{tabular}{l|ccc|cc}
    \toprule
    \multirow{2}{*}{Task}
    & \multicolumn{3}{c|}{BERTScore-F1}
    & \multicolumn{2}{c}{Error decomposition} \\
    \cmidrule(lr){2-4}
    \cmidrule(lr){5-6}
    & \textit{Correct}
    & \textit{Shuffle}
    & \textit{Generic}
    & $\mathrm{S}^{+}/\mathrm{B}^{-}$
    & $\mathrm{S}^{-}/\mathrm{B}^{+}$ \\
    \midrule

    RGB
    & 94\%
    & 69\%
    & 79\%
    & 4.26
    & 10.06 \\

    RGB-D
    & 93\%
    & 73\%
    & 84\%
    & 3.13
    & 9.87 \\

    RGB-T
    & 94\%
    & 71\%
    & 86\%
    & 5.44
    & 16.41 \\
    \bottomrule
  \end{tabular}
\end{table}

\subsection{Evaluation of MLLMs Grounding and Semantics}

Figure~\ref{fig_intro} compares MLLM grounding on SaliLLM with fully supervised methods, while Table~\ref{tab:phrase_recognition} evaluates phrase-level recognition and localization. Across the three modalities, correctly paired phrases achieve an average BERTScore-F1 \cite{zhang2019bertscore} of 93.67\%, exceeding the shuffled-GT and generic-phrase baselines by 23\% and 11\%, respectively. After one-to-one entity matching, semantic-correct/box-wrong ($\mathrm{S}^{+}/\mathrm{B}^{-}$) and semantic-wrong/box-correct ($\mathrm{S}^{-}/\mathrm{B}^{+}$) cases account for 4.28\% and 12.11\% on average. These results demonstrate strong zero-shot salient-entity perception of MLLMs, while revealing that accurate localization does not necessarily imply correct semantic grounding.

\subsection{Dissecting Salient Perception in MLLMs}
\label{sec:diagnostic_analysis}
To uncover the root causes of these capability failures, we conduct a systematic diagnostic analysis of Qwen2.5~\cite{bai2025qwen25}, Qwen3~\cite{bai2025qwen3}, and InternVL3.5~\cite{wang2025internvl3}. Specifically, we investigate both their ability and characteristic to perceive salient objects and the internal mechanisms underlying their predictions. At selected layers, image-token states are replaced by their within-layer mean, and localization change is measured against the unmodified deterministic decode. Comparisons within each model share images and prompts, and layer indices are normalized by model depth.

\subsubsection{Evidence I: When Do MLLMs Decide?}

The top row of Fig.~\ref{fig_analysis} characterizes when salient evidence stabilizes and which layers support its formation. To quantify this process, we introduce \textit{Layer-wise Grounding Saturation Depth} (LGSD), defined as the normalized decoding depth at which the grounding representation first reaches 90\% of its final probing strength \cite{geva2022transformer}. A higher LGSD indicates that the model requires deeper feature integration before forming a stable perception. Fig.~\ref{fig_analysis} \textbf{\textit{(A1)--(A3)}} show that the proportion of late-stabilizing cases in Qwen3-VL increases from 9\% for simple objects to 31\% for complex objects. The same panels further reveal that reducing object scale or increasing shape complexity raises LGSD by approximately 10\%, suggesting that visually challenging objects require additional processing depth. Fig.~\ref{fig_analysis} \textit{\textbf{(A4)}} confirms this trend more directly: the smaller the object area, the deeper the layer at which the model's perception stabilizes. Fig.~\ref{fig_analysis} \textit{\textbf{(B)}} complements these probing results with layer-wise ablations, showing that grounding relies primarily on visual tokens from the early and middle layers, while complex objects retain causal support into later layers. \textbf{\textit{Evidence I:}} These results reveal a hierarchical mechanism that dynamically allocates computational depth according to the perceptual complexity of an object. Large and simple objects can trigger stable perception early through concentrated visual evidence, whereas small and structurally complex objects require deeper and more sustained visual integration.

\subsubsection{Evidence II: How Do MLLMs Find the Object?}
The bottom row reveals how semantic selection is converted into spatial localization, and where this conversion breaks down. In Fig.~\ref{fig_analysis} (\textit{\textbf{C1}}) and (\textbf{\textit{C3}}), the target enrichment of Qwen3-VL increases from $1.6$ at the \textit{phrase} stage to $1.8$ at the \textit{box} stage. InternVL3.5 exhibits stronger enrichment at the \textit{phrase} stage, reaching $2.5$, but this value drops to $1.6$ at the \textit{box} stage, yielding a final IoU of only $0.35$. Fig.~\ref{fig_analysis} (\textit{\textbf{C2}}) further shows that target occlusion, incorrect phrases, and disrupted foreground background evidence induce localization changes of $0.61$, $0.56$, and $0.55$, respectively, whereas background occlusion has only a marginal effect of $-0.034$. These results indicate that localization is jointly driven by target specific visual evidence and linguistic conditioning. As shown in Fig.~\ref{fig_analysis} (\textbf{\textit{D}}), complex objects also shift the causal center of mass toward deeper layers by $0.083$ for Qwen3-VL and $0.135$ for InternVL3.5, while distributing supporting evidence across a broader range of layers. Nevertheless, Fig.~\ref{fig_analysis} (\textbf{\textit{E}}) shows that the predicted boxes achieve 97\% coverage and an enclosing box IoU of 0.89, yet attain a purity of only $0.48$ and an area ratio of 2.86. \textbf{\textit{Evidence II:}} This reveals a critical capability decoupling: MLLMs can already select and coarsely localize salient entities, but strong semantic evidence does not automatically consolidate into accurate object extents.

\subsubsection{Mechanistic Implication: What Is Still Missing?}
The two lines of evidence show that MLLMs can extract target specific visual cues, identify salient entities, and localize them coarsely, yet still lack a reliable mechanism for semantic spatial closure. They fail to consolidate local discriminative evidence into foreground regions that are complete in extent, coherent along object boundaries, and aligned with the observer protocol. The central bottleneck of zero-shot SOD is therefore protocol conditioned foreground organization: determining the appropriate granularity, extent, and inclusion relationships of selected entities in the final mask. This diagnosis directly motivates the design of our method.

\begin{figure*}[!t]
    \centering
    \captionsetup{skip=5pt}
    \includegraphics[width=1\linewidth]{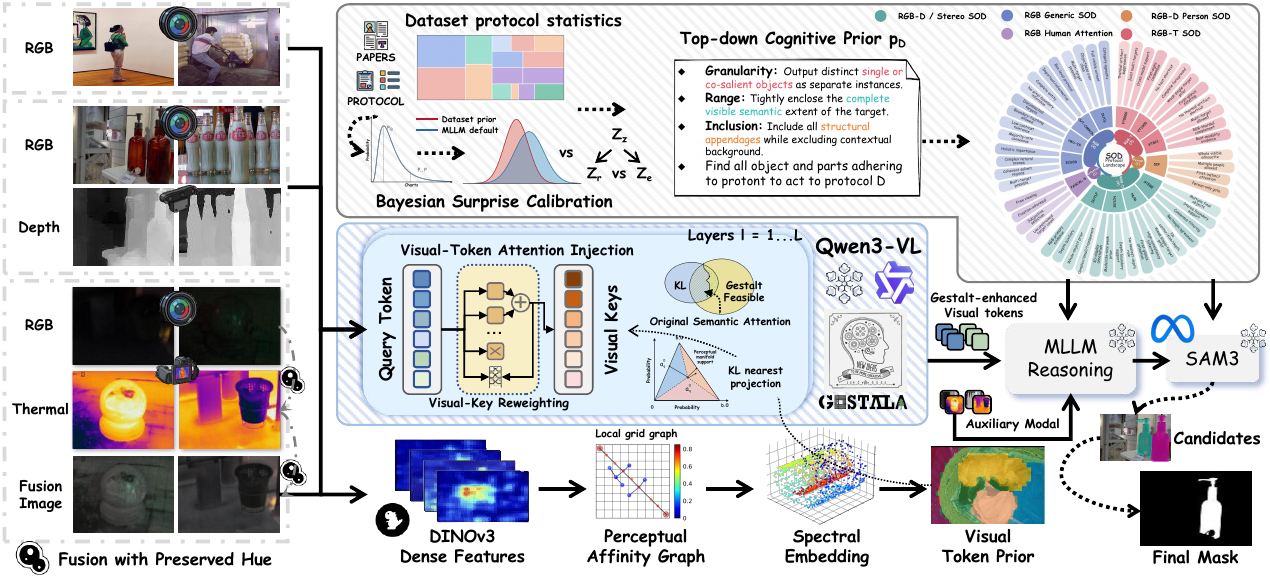}
    \caption{FOCUS: protocol-conditioned foreground organization through collaborative top-down and bottom-up attention.}
    \label{fig_method}
    \vspace{-1mm}
\end{figure*}

\section{Method}
\label{sec:Method}

\subsection{Framework Overview}
\label{sec:overview}
As shown in Fig.~\ref{fig_method}, FOCUS maps $X=(I,I^a)$, where $I^a$ is optional depth or thermal input, to a mask. A top-down observer protocol $p_D$ constrains foreground granularity, extent, and inclusion, while DINOv3 \cite{simeoni2025dinov3} graphs and soft charts produce bottom-up visual support $\mathbf s_X$. Their joint attention yields a phrase-box $h=(\hat t,\hat B)$; SAM3 generates mask candidates, and Protocol-Conditioned Bayesian Surprise (PCBS) selects $\hat S$. Multiple phrase-boxes are processed independently and merged by pixel-wise union. We view this process as a protocol-conditioned Gestalt projection of the original attention $\mathbf a^0(p_D)$:
\begin{equation}
\mathbf a^*=\arg\min_{\mathbf a\in\mathcal M_X\cap\mathcal P_D}
D_{\rm KL}\!\left(\mathbf a\,\|\,\mathbf a^0(p_D)\right).
\label{eq:focus_projection}
\end{equation}
$\mathcal M_X$ enforces graph smoothness and ground separation, whereas $\mathcal P_D$ constrains granularity, extent, and inclusion. We approximate Eq.~\eqref{eq:focus_projection} with $p_D$ and one visual-token update.

\subsection{Top-Down Observer-Protocol Calibration}
\label{sec:cognitive_prior}

$z=(z_g,z_r,z_e)$ denotes protocol granularity, extent, and inclusion. Let $\mathcal R_D$ contain only the original dataset paper and public acquisition or annotation descriptions. Fixed verbalizers map documented statements to categorical choices (single/group foreground, tight/broad extent, and inclusion/exclusion of attached or contextual regions), yielding $\pi_D^u=\pi_D(z_u\mid\mathcal R_D)$. A dimension absent from the public description receives a uniform prior and adds no dataset-specific rule. The neutral instruction $p_0$ induces the dataset-agnostic prior $q_0^u=q_\theta(z_u\mid p_0)$. Their Bayesian surprise and normalized weight are
\begin{equation}
\begin{aligned}
b_D^u&=D_{\rm KL}\!\left(q_0^u\,\|\,\pi_D^u\right),\\
\rho_D^u&=\frac{b_D^u}{\sum_v b_D^v+\epsilon},\quad \epsilon>0.
\end{aligned}
\label{eq:bayesian_surprise}
\end{equation}
The deterministic template $p_D=\mathcal T(\mathcal R_D,\boldsymbol\rho_D)$ strengthens dimensions with larger mismatches. Each $p_D$ is generated once from public documents and frozen before image inference; with no documented protocol, all dimensions remain uniform and the method falls back to $p_0$. Protocol construction reads no benchmark image, while PCBS uses only the current image and its generated candidates. Neither stage accesses test masks, object counts, area distributions, or evaluation outputs; all ground truth is loaded only after predictions are finalized. The same protocol variables are reused by PCBS during mask selection.

\subsection{Bottom-Up Perceptual-Manifold Injection}
\label{sec:local_cue_manifold}

DINOv3 groups patches that may form the same object. From normalized features $\mathbf f_m$, positions $\mathbf r_m$, and low-frequency graph coordinates $\mathbf u_m$, we construct
\begin{equation}
\begin{aligned}
A_{mn}
&=\exp\!\left[
\tau_f(\mathbf f_m^\top\mathbf f_n-1)
-\frac{\|\mathbf r_m-\mathbf r_n\|_2^2}{2\sigma_p^2}
\right]\mathbb I_{mn},\\
q_{mc}
&=\operatorname{softmax}_{c}\!\left(
-\frac{\|\mathbf u_m-\boldsymbol\mu_c\|_2^2}{\sigma_a^2}
\right),\\
\mathcal K_D&=\operatorname{TopK}(o;K_D),
s_m=\operatorname{norm}_{[0,1]}\!\left[
\sum_{c\in\mathcal K_D}q_{mc}\omega(o_c)\right].
\end{aligned}
\label{eq:affinity}
\end{equation}
$\mathbb I_{mn}$ is symmetrized $k$-NN connectivity, $q_{mc}$ is membership to chart center $\boldsymbol\mu_c$, and $\tau_f,\sigma_p,\sigma_a$ are feature and chart scales. The graph Laplacian $L_X=\operatorname{diag}(A\mathbf1)-A$ limits diffusion across weak affinities; spectral soft charts preserve global grouping and boundary uncertainty. Chart score $o_c$ combines centrality, boundary contact, and optional auxiliary contrast. The top $K_D$ charts yield $s_m$, which is resampled to the MLLM token grid; exact definitions are provided in the supplement. After resampling $s_m$, let
$\mathbf a_q^0=\operatorname{softmax}_{\mathcal K_q}(\boldsymbol\ell_q)$
be the protocol-conditioned attention and
$g_j=\mathbb I[j\in\mathcal V]\log(1+s_j)$.
FOCUS exactly tilts this attention toward coherent visual keys by
$(1+s_j)$ through
\begin{equation}
\begin{aligned}
\mathbf a_q^*
&=\arg\min_{\mathbf a\in\Delta(\mathcal K_q)}
\left[D_{\rm KL}(\mathbf a\|\mathbf a_q^0)
-\langle\mathbf a,\mathbf g\rangle\right] \\
&=\operatorname{softmax}_{\mathcal K_q}
(\boldsymbol\ell_q+\mathbf g).
\end{aligned}
\label{eq:attention_injection}
\end{equation}

\subsection{Protocol-Consistent Masks}
\label{sec:mask_realization}

Depth supplies auxiliary contrast for RGB-D, while frozen PHFuse~\cite{zhuang2025phfuse} fuses aligned visible and thermal inputs for RGB-T. Given $h^*=f_\theta(X;p_D,\mathbf s_X)$, SAM3~\cite{carion2025sam} returns $\mathcal Q(h^*)=\{(m_j,s_j^{\rm sam})\}_{j=1}^{J}$. A fixed rubric produces $\mathbf v_X=(k_X,\ell_X,c_X,d_X)$ for surprise, protocol likelihood, completeness, and distractor risk. Each mask is represented by $\mathbf r_D(m)=[B,C,G,A_{\rm aux},-L,-F]^\top$, encoding box agreement, support recall, chart coherence, auxiliary boundary agreement, leakage, and fragmentation. PCBS computes $\eta_X=\boldsymbol\beta^\top(k_X,\ell_X,c_X,-d_X)$ and uses the fixed per-image gate $\lambda_X=\lambda_{\min}+(\lambda_{\max}-\lambda_{\min})\sigma(\eta_X)$ to select
\begin{equation}
\begin{aligned}
j^*&=\arg\max_{1\leq j\leq J}\Bigl[
(1-\lambda_X)s_j^{\rm sam}\\
&\quad+\lambda_X\operatorname{clip}_{[0,1]}
\langle\Psi_D(\mathbf v_X),\mathbf r_D(m_j)\rangle\Bigr].
\end{aligned}
\label{eq:pcbs_reranking}
\end{equation}
The final mask is $\hat S=m_{j^*}$. Here, $\sigma$ is the logistic function and $\Psi_D$ deterministically maps protocol dimensions to candidate criteria. All coefficients, candidate budgets, and manifold settings are shared across datasets and fixed before evaluation. PCBS only reranks masks generated for the selected entity and never changes entity identity.

\begin{table*}[t]
  \centering
  \small
  \caption{Quantitative comparison on RGB datasets. The best and second-best results are highlighted in bold and underlined.}
  \label{RGB_SOTA}
  \renewcommand{\arraystretch}{0.7}
  \setlength\tabcolsep{2.5pt}
  \begin{tabular}{l|c|cccc|cccc|cccc|cccc|cccc}
    \toprule
    \multirow{2}{*}{Methods}
    & \multirow{2}{*}{Year}
    & \multicolumn{4}{c}{DUTS}
    & \multicolumn{4}{c}{DUT-O}
    & \multicolumn{4}{c}{HKU-IS}
    & \multicolumn{4}{c}{PASCAL-S}
    & \multicolumn{4}{c}{ECSSD} \\
    \cmidrule(lr){3-6}
    \cmidrule(lr){7-10}
    \cmidrule(lr){11-14}
    \cmidrule(lr){15-18}
    \cmidrule(lr){19-22}
    &
    & \textit{E}\textsubscript{\textit{m}}$\uparrow$
    & \textit{F}\textsubscript{\textit{m}}$\uparrow$
    & \textit{S}\textsubscript{\textit{m}}$\uparrow$
    & \textit{M}$\downarrow$
    & \textit{E}\textsubscript{\textit{m}}$\uparrow$
    & \textit{F}\textsubscript{\textit{m}}$\uparrow$
    & \textit{S}\textsubscript{\textit{m}}$\uparrow$
    & \textit{M}$\downarrow$
    & \textit{E}\textsubscript{\textit{m}}$\uparrow$
    & \textit{F}\textsubscript{\textit{m}}$\uparrow$
    & \textit{S}\textsubscript{\textit{m}}$\uparrow$
    & \textit{M}$\downarrow$
    & \textit{E}\textsubscript{\textit{m}}$\uparrow$
    & \textit{F}\textsubscript{\textit{m}}$\uparrow$
    & \textit{S}\textsubscript{\textit{m}}$\uparrow$
    & \textit{M}$\downarrow$
    & \textit{E}\textsubscript{\textit{m}}$\uparrow$
    & \textit{F}\textsubscript{\textit{m}}$\uparrow$
    & \textit{S}\textsubscript{\textit{m}}$\uparrow$
    & \textit{M}$\downarrow$ \\
    \midrule

    \multicolumn{22}{c}{\textbf{\textit{Fully supervised}}} \\
    \midrule

    VST++
    & 2024
    & .947 & .897 & .909 & .029
    & .890 & .813 & .859 & .050
    & \textbf{.969} & .941 & .932 & .025
    & .901 & .859 & .880 & .062
    & \underline{.969} & .951 & \underline{.939} & .027 \\

    PAM
    & 2025
    & .944 & .901 & .903 & .029
    & .904 & .831 & .859 & .048
    & \underline{.965} & \underline{.944} & \underline{.933} & \underline{.023}
    & \underline{.917} & \underline{.895} & \underline{.887} & .058
    & .966 & \underline{.957} & .937 & .026 \\

    RMFDNet
    & 2025
    & .927 & .903 & .897 & .034
    & .869 & .832 & .845 & .052
    & .953 & \underline{.944} & .923 & .028
    & .901 & .884 & .865 & .060
    & .947 & .953 & .927 & .033 \\

    RMFNet
    & 2026
    & \underline{.949} & \underline{.914} & \underline{.917} & \underline{.025}
    & .906 & .829 & \underline{.873} & \underline{.042}
    & \underline{.965} & \textbf{.950} & \textbf{.936} & \textbf{.022}
    & \underline{.917} & .888 & .882 & \underline{.045}
    & .964 & .948 & .938 & .024 \\

    Saliency-R1
    & 2026
    & -- & -- & -- & --
    & \underline{.907} & \underline{.833} & .864 & .049
    & -- & -- & -- & --
    & .915 & .891 & \underline{.887} & .049
    & .968 & .952 & .935 & \textbf{.022} \\

    \midrule
    \multicolumn{22}{c}{\textbf{\textit{Weakly supervised}}} \\
    \midrule

    NSAL
    & 2023
    & .850 & .738 & .781 & .073
    & .802 & .656 & .745 & .088
    & .923 & .867 & .854 & .051
    & .826 & .759 & .767 & .110
    & .889 & .857 & .834 & .078 \\

    WBNet
    & 2024
    & .915 & .852 & .876 & .038
    & .894 & .804 & .855 & .048
    & .958 & .920 & .913 & .029
    & .872 & .840 & .851 & .066
    & .918 & .929 & .937 & .032 \\

    \midrule
    \multicolumn{22}{c}{\textbf{\textit{Self-supervised}}} \\
    \midrule

    CutLER
    & 2023
    & .767 & .660 & .737 & .069
    & .717 & .590 & .696 & .159
    & .886 & .837 & .844 & .073
    & .834 & .764 & .786 & .109
    & .887 & .850 & .846 & .077 \\

    3SD
    & 2024
    & .798 & .755 & .812 & .087
    & .790 & .720 & .798 & .094
    & .906 & .877 & .877 & .068
    & .812 & .791 & .807 & .117
    & .891 & .898 & .885 & .077 \\

    CSNet
    & 2025
    & .915 & .852 & .853 & .043
    & .846 & .769 & .794 & .068
    & .949 & .926 & .901 & .031
    & .885 & .834 & .831 & .071
    & .922 & .912 & .886 & .048 \\

    \midrule
    \multicolumn{22}{c}{\textbf{\textit{Zero-shot}}} \\
    \midrule

    GPT-4o
    & 2024
    & .825 & .811 & .812 & .092
    & .788 & .742 & .753 & .105
    & .861 & .857 & .845 & .057
    & .736 & .715 & .718 & .152
    & .727 & .718 & .707 & .203 \\

    Gemini-2.5
    & 2025
    & .898 & .870 & .860 & .060
    & .851 & .781 & .807 & .077
    & .925 & .914 & .881 & .038
    & .901 & .875 & .861 & .055
    & .929 & .894 & .898 & .047 \\

    Qwen2.5-VL
    & 2025
    & .729 & .701 & .693 & .115
    & .637 & .601 & .613 & .215
    & .720 & .713 & .697 & .161
    & .588 & .559 & .553 & .271
    & .531 & .510 & .518 & .372 \\
    
    Qwen3-VL
    & 2025
    & .898 & .853 & .868 & .064
    & .796 & .719 & .768 & .133
    & .946 & .919 & .908 & .036
    & .881 & .840 & .844 & .078
    & .938 & .931 & .910 & .038 \\


    \textit{\textbf{FOCUS (Ours)}}
    & --
    & \textbf{.961} & \textbf{.928} & \textbf{.927} & \textbf{.024}
    & \textbf{.924} & \textbf{.858} & \textbf{.881} & \textbf{.036}
    & \textbf{.969} & \underline{.944} & .928 & \underline{.023}
    & \textbf{.944} & \textbf{.907} & \textbf{.903} & \textbf{.037}
    & \textbf{.973} & \textbf{.964} & \textbf{.940} & \underline{.023} \\
    \bottomrule
  \end{tabular}
\end{table*}

\begin{table*}[t!]
  \centering
  \small
   \caption{Quantitative comparison on RGB-D datasets. The best and second-best results are highlighted in bold and underlined.}
  \label{RGBD_SOTA}
  \renewcommand{\arraystretch}{0.7}
  \setlength\tabcolsep{2.5pt}
  \begin{tabular}{l|c|cccc|cccc|cccc|cccc|cccc}
    \toprule
    \multirow{2}{*}{Methods}
    & \multirow{2}{*}{Year}
    & \multicolumn{4}{c}{SIP}
    & \multicolumn{4}{c}{STERE}
    & \multicolumn{4}{c}{DUTLF-D}
    & \multicolumn{4}{c}{NLPR}
    & \multicolumn{4}{c}{NJUD} \\
    \cmidrule(lr){3-6}
    \cmidrule(lr){7-10}
    \cmidrule(lr){11-14}
    \cmidrule(lr){15-18}
    \cmidrule(lr){19-22}
    &
    & \textit{E}\textsubscript{\textit{m}}$\uparrow$
    & \textit{F}\textsubscript{\textit{m}}$\uparrow$
    & \textit{S}\textsubscript{\textit{m}}$\uparrow$
    & \textit{M}$\downarrow$
    & \textit{E}\textsubscript{\textit{m}}$\uparrow$
    & \textit{F}\textsubscript{\textit{m}}$\uparrow$
    & \textit{S}\textsubscript{\textit{m}}$\uparrow$
    & \textit{M}$\downarrow$
    & \textit{E}\textsubscript{\textit{m}}$\uparrow$
    & \textit{F}\textsubscript{\textit{m}}$\uparrow$
    & \textit{S}\textsubscript{\textit{m}}$\uparrow$
    & \textit{M}$\downarrow$
    & \textit{E}\textsubscript{\textit{m}}$\uparrow$
    & \textit{F}\textsubscript{\textit{m}}$\uparrow$
    & \textit{S}\textsubscript{\textit{m}}$\uparrow$
    & \textit{M}$\downarrow$
    & \textit{E}\textsubscript{\textit{m}}$\uparrow$
    & \textit{F}\textsubscript{\textit{m}}$\uparrow$
    & \textit{S}\textsubscript{\textit{m}}$\uparrow$
    & \textit{M}$\downarrow$ \\
    \midrule

    \multicolumn{22}{c}{\textbf{\textit{Fully supervised}}} \\
    \midrule



    HENet
    & 2025
    & .946 & .916 & \underline{.914} & \underline{.032}
    & .939 & .914 & \underline{.926} & \underline{.029}
    & \underline{.968} & .946 & \underline{.946} & \underline{.022}
    & .970 & .913 & \textbf{.942} & \underline{.016}
    & .906 & .891 & .900 & .049 \\
    
    MambaSOD
    & 2025
    & .940 & .914 & .904 & .040
    & \underline{.955} & .920 & .924 & .031
    & .967 & \underline{.947} & .942 & .024
    & \textbf{.973} & \underline{.934} & \underline{.941} & .017
    & \textbf{.963} & \underline{.937} & \textbf{.934} & \textbf{.027} \\
    
    STENet
    & 2026
    & .945 & \underline{.920} & .908 & .036
    & .949 & .910 & .911 & .034
    & .966 & .944 & .935 & .026
    & .964 & .929 & .933 & .020
    & .953 & .930 & .923 & .029 \\
    
    HMaT-D
    & 2026
    & \underline{.947} & .914 & .907 & .038
    & \underline{.955} & \underline{.921} & .920 & .031
    & .967 & .946 & .941 & .026
    & .965 & .923 & .934 & .023
    & .958 & .929 & .927 & .032 \\

    \midrule
    \multicolumn{22}{c}{\textbf{\textit{Weakly supervised}}} \\
    \midrule

    MIRV
    & 2024
    & .924 & .863 & .876 & .049
    & .934 & .873 & .891 & .040
    & .871 & .811 & .834 & .076
    & .953 & .895 & .914 & .025
    & .929 & .880 & .890 & .046 \\

    Ding et al.
    & 2025
    & .915 & .867 & .872 & .054
    & .927 & .892 & .902 & .038
    & .958 & .934 & .930 & .025
    & .958 & .899 & .928 & .021
    & .919 & .912 & .912 & .038 \\

    \midrule
    \multicolumn{22}{c}{\textbf{\textit{Self-supervised}}} \\
    \midrule
    A2Sv3
    & 2024
    & .932 & .889 & .881 & .044
    & .941 & .895 & .892 & .040
    & .908 & .869 & .874 & .054
    & .949 & .892 & .905 & .028
    & .924 & .882 & .881 & .049 \\
    
    Qi et al.
    & 2025
    & .929 & .909 & .885 & .045
    & .927 & .899 & .886 & .046
    & .898 & .859 & .862 & .081
    & .936 & .915 & .921 & .046
    & .927 & .899 & .907 & .040 \\

    Gao et al.
    & 2025
    & .924 & .894 & .885 & .051
    & .942 & .900 & .906 & .040
    & .912 & .871 & .879 & .052
    & .958 & .910 & .921 & .026
    & .900 & .853 & .867 & .054 \\

    \midrule
    \multicolumn{22}{c}{\textbf{\textit{Zero-shot}}} \\
    \midrule
    GPT-4o
    & 2024
    & .836 & .803 & .799 & .132
    & .784 & .759 & .747 & .185
    & .795 & .779 & .773 & .154
    & .836 & .805 & .813 & .098
    & .779 & .750 & .742 & .163 \\
    
    Gemini-2.5
    & 2025
    & .920 & .896 & .892 & .051
    & .899 & .875 & .863 & .050
    & .891 & .874 & .868 & .058
    & .924 & .872 & .899 & .049
    & .911 & .885 & .889 & .066 \\

    Qwen2.5-VL
    & 2025
    & .631 & .612 & .603 & .298
    & .554 & .516 & .511 & .363
    & .618 & .598 & .592 & .311
    & .633 & .603 & .617 & .287
    & .682 & .658 & .648 & .269 \\

    Qwen3-VL
    & 2025
    & .915 & .883 & .888 & .055
    & .918 & .889 & .888 & .052
    & .882 & .870 & .863 & .087
    & .929 & .888 & .897 & .039
    & .926 & .899 & .893 & .045 \\


    \textbf{\textit{FOCUS (Ours)}}
    & --
    & \textbf{.989} & \textbf{.980} & \textbf{.965} & \textbf{.010}
    & \textbf{.960} & \textbf{.933} & \textbf{.929} & \textbf{.027}
    & \textbf{.971} & \textbf{.962} & \textbf{.947} & \textbf{.021}
    & \underline{.972} & \textbf{.939} & .940 & \textbf{.015}
    & \underline{.961} & \textbf{.947} & \underline{.929} & \underline{.028} \\

    \bottomrule
  \end{tabular}
\end{table*}

\begin{table}[t]
  \centering
  \small
    \caption{Quantitative comparison on RGB-T SOD datasets.}
  \label{RGBT_SOTA}
  \renewcommand{\arraystretch}{0.7}
  \setlength\tabcolsep{2.6pt}
  \begin{tabular}{l|ccc|ccc|ccc}
    \toprule
    \multirow{2}{*}{Methods\textsubscript{Year}}
    & \multicolumn{3}{c}{VT821}
    & \multicolumn{3}{c}{VT5000}
    & \multicolumn{3}{c}{VT1000} \\
    \cmidrule(lr){2-4}
    \cmidrule(lr){5-7}
    \cmidrule(lr){8-10}
    & \textit{E}\textsubscript{\textit{m}}$\uparrow$
    & \textit{S}\textsubscript{\textit{m}}$\uparrow$
    & \textit{M}$\downarrow$
    & \textit{E}\textsubscript{\textit{m}}$\uparrow$
    & \textit{S}\textsubscript{\textit{m}}$\uparrow$
    & \textit{M}$\downarrow$
    & \textit{E}\textsubscript{\textit{m}}$\uparrow$
    & \textit{S}\textsubscript{\textit{m}}$\uparrow$
    & \textit{M}$\downarrow$ \\
    \midrule

    \multicolumn{10}{c}{\textbf{\textit{Fully supervised}}} \\
    \midrule

    DiMSOD\textsubscript{25}
    & .949 & .923 & .025
    & .959 & .921 & .029
    & .935 & \underline{.950} & .020 \\

    ConTriNet\textsubscript{25}
    & .940 & .915 & .022
    & .956 & .923 & \underline{.020}
    & .953 & .941 & \underline{.015} \\

    DFNet\textsubscript{26}
    & .940 & \underline{.926} & .022
    & .958 & \underline{.930} & \underline{.020}
    & .955 & \underline{.950} & \textbf{.013} \\

    SAMSOD\textsubscript{26}
    & \underline{.956} & .925 & \underline{.020}
    & \underline{.964} & .923 & .021
    & \underline{.976} & .942 & \underline{.015} \\

    GaCNet\textsubscript{26}
    & .947 & \textbf{.929} & .021
    & .944 & .919 & \textbf{.019}
    & .966 & .948 & .019 \\

    \midrule
    \multicolumn{10}{c}{\textbf{\textit{Weakly supervised}}} \\
    \midrule

    PCCFU\textsubscript{25}
    & .946 & .912 & .024
    & .948 & .907 & .026
    & .972 & .943 & .016 \\

    SSFam\textsubscript{25}
    & .943 & .919 & .023
    & .951 & .914 & .026
    & .973 & .943 & \underline{.015} \\

    \midrule
    \multicolumn{10}{c}{\textbf{\textit{Self-supervised}}} \\
    \midrule

    A2Sv3\textsubscript{24}
    & .922 & .882 & .041
    & .924 & .872 & .038
    & .959 & .923 & .023 \\

    \midrule
    \multicolumn{10}{c}{\textbf{\textit{Zero-shot}}} \\
    \midrule

    GPT-4o\textsubscript{24}
    & .836 & .814 & .091
    & .824 & .789 & .077
    & .844 & .812 & .093 \\

    Gemini-2.5\textsubscript{25}
    & .917 & .886 & .053
    & .891 & .853 & .042
    & .936 & .892 & .057 \\

    Qwen2.5-VL\textsubscript{25}
    & .832 & .798 & .112
    & .812 & .776 & .108
    & .829 & .801 & .097 \\

    Qwen3-VL\textsubscript{25}
    & .914 & .888 & .047
    & .898 & .866 & .059
    & .933 & .905 & .042 \\

    \textbf{\textit{FOCUS (Ours)}}
    & \textbf{.959} & \textbf{.929} & \textbf{.019}
    & \textbf{.966} & \textbf{.932} & \textbf{.019}
    & \textbf{.981} & \textbf{.952} & \textbf{.013} \\

    \bottomrule
  \end{tabular}
\end{table}

\begin{table}[!t]
  \centering
  \small
   \caption{Ablations of key FOCUS components. Inst./Obs. denote native/observer prompts; Man./Prop. denote manifold/propagation. Results are $S_m/\mathrm{MAE}$.}
  \label{tab:focus_ablation}
  \renewcommand{\arraystretch}{0.90}
  \setlength{\tabcolsep}{1.35pt}
  \begin{tabular}{@{}lccccc ccc@{}}
    \toprule
    \multirow{2}{*}{Vari.}
      & \multicolumn{2}{c}{Top-down}
      & \multicolumn{2}{c}{Bottom-up}
      & \multirow{2}{*}{PCBS}
      & \multicolumn{3}{c}{Datasets} \\
    \cmidrule(lr){2-3}\cmidrule(lr){4-5}\cmidrule(l){7-9}
      & Prompt & BS & Man. & Prop. &
      & DUTS & DUTF & VT5K \\
    \midrule
    V1
      & Inst. & -- & -- & -- & --
      & .868/.064 & .863/.087 & .866/.059 \\
    V2
      & Obs. & -- & -- & -- & --
      & .900/.032 & .926/.037 & .903/.035 \\
    V3
      & $p_D$ & \ding{51} & -- & -- & --
      & .907/.028 & .928/.035 & .909/.031 \\
    V4
      & $p_D$ & \ding{51} & \ding{51} & \ding{51} & --
      & .921/.024 & .935/.029 & .921/.025 \\
    \textbf{Ours}
      & $p_D$ & \ding{51} & \ding{51} & \ding{51} & \ding{51}
      & \textbf{.927}/\textbf{.020} & \textbf{.947/.021} & \textbf{.932/.019} \\
    \bottomrule
  \end{tabular}
\end{table}

\section{Experiments}
\label{sec:Experiments}

\subsection{Datasets and Metrics}

For \colorbox{gray!20}{RGB SOD}, we evaluate FOCUS on five commonly used benchmark datasets, i.e., \textbf{DUTS} \cite{wang2017learning}, \textbf{DUT-O} \cite{yang2013saliency}, \textbf{HKU-IS} \cite{li2015visual}, \textbf{PASCAL-S} \cite{li2014secrets} and \textbf{ECSSD} \cite{yan2013hierarchical}. 
As for \colorbox{gray!20}{RGB-D SOD}, we use five benchmark datasets, including \textbf{NJUD} \cite{ju2014depth}, \textbf{NLPR} \cite{peng2014rgbd}, \textbf{SIP} \cite{fan2020rethinking}, \textbf{STERE} \cite{niu2012leveraging} and \textbf{DUTLF-D} \cite{piao2019depth}.
Regarding \colorbox{gray!20}{RGB-T SOD}, we employ three benchmark datasets: \textbf{VT821} \cite{wang2018rgb}, \textbf{VT1000} \cite{tu2019rgb} and \textbf{VT5000} \cite{tu2022rgbt}. 
We adopt four metrics to evaluate model performance, i.e., structure-measure ($S_m$) \cite{fan2017structure}, maximum F-measure ($F_m$) \cite{achanta2009frequency}, maximum enhanced-alignment measure ($E_m$) \cite{fan2018enhanced}, and mean absolute error (\textit{M}).

\subsection{Experimental Setup}

\subsubsection{Baselines.} For RGB SOD, we compare against supervised methods, including VST++, PAM~\cite{khan2025pyramidal}, RMFDNet~\cite{zhou2025rmfdnet}, RMFNet~\cite{zhang2026salient}, and Saliency-R1; weakly supervised methods, including NSAL~\cite{piao2022noise} and WBNet~\cite{wang2024wbnet}; and self-supervised methods, including CutLER~\cite{wang2023cut}, 3SD~\cite{yasarla20243sd}, and CSNet~\cite{guan2025contrastive}. For RGB-D, the compared methods include HENet~\cite{gao2024highly}, MambaSOD~\cite{zhan2025mambasod}, STENet~\cite{chen2026stenet}, HMaT-D~\cite{tan2026multi}, MIRV~\cite{li2023mutual}, Ding et al.~\cite{ding2025cross}, A2Sv3~\cite{yuan2024unified}, Qi et al.~\cite{qi2024masked}, and Gao et al.~\cite{gao2025self}. For RGB-T, we compare against DiMSOD~\cite{zhang2025dimsod}, ConTriNet~\cite{tang2024divide}, DFNet~\cite{hou2025hypsam}, SAMSOD~\cite{liu2026samsod}, GaCNet~\cite{hou2026empirical}, PCCFU~\cite{zhai2025weakly}, SSFam~\cite{liu2025ssfam}, and A2Sv3~\cite{yuan2024unified}. We additionally evaluate GPT-4o, Gemini-2.5, Qwen2.5-VL, and Qwen3-VL under the same realization pipeline with SAM3. 

\subsubsection{Implementation Details.} FOCUS uses Qwen3-VL-8B-Instruct, DINOv3 ViT-L/16, and SAM3 with BF16 and deterministic decoding. All manifold, graph, and candidate-generation settings are fixed globally before evaluation and shared across datasets; the exact configuration is provided in the supplementary material.
All experiments are implemented in PyTorch and conducted on 4 NVIDIA 4090 GPUs. The inference configuration is kept fixed throughout the process across all datasets.

\subsection{Performance Analysis}
\label{sec:sota_comparison}
Tables~\ref{RGB_SOTA}--\ref{RGBT_SOTA} show that FOCUS achieves consistently strong performance across RGB, RGB-D, and RGB-T benchmarks. It substantially outperforms self supervised and weakly supervised methods, remains competitive with fully supervised models tailored to specific modalities, and surpasses MLLM baselines. FOCUS reduces mean absolute error by 11\%, 34\%, and 48\% compared with fully, weakly, and self-supervised methods, respectively.
These results show that model scale alone is insufficient to bridge the gap between semantic localization and  segmentation. All results are obtained with the same frozen foundation models, demonstrating a unified training-free solution for zero-shot SOD.

\subsection{Efficiency and Resource Overhead}
\label{sec:efficiency}

Under identical hardware and inference settings, FOCUS processes each image in 2.19s with 18.14GB peak GPU memory. In comparison, ProMaC~\cite{hu2024leveraging} requires 130.49s and 32.85GB, while RDVP-MSD~\cite{yin2025stepwise} requires 18.05s and 31.51GB. FOCUS is therefore $59.6\times$ and $8.2\times$ faster, while reducing peak memory by $44.8\%$ and $42.4\%$, respectively. Consequently, these results demonstrate that FOCUS combines strong zero-shot SOD performance with substantially lower inference overhead.

\subsection{Ablation Study}
\label{sec:ablation}

Table~\ref{tab:focus_ablation} uses a cumulative, equal-budget design: all variants share Qwen3-VL, SAM3, image resolution, decoding, and candidate generation. V1 is the native-instruction Qwen3-VL+SAM3 baseline; V2 changes only to the documented observer prompt; V3 adds Bayesian-surprise weighting; V4 adds the complete manifold pathway; FOCUS adds PCBS. Experiments are conducted on three datasets: DUTS (DUTS-TE), DUTF (DUTLF-D), and VT5K (VT5000).
The observer protocol provides the largest initial gain, while Bayesian-surprise calibration and bottom-up organization improve both metrics across all modalities. PCBS further improves $S_m$ and MAE on all three datasets. These controlled comparisons isolate the contributions of protocol calibration, perceptual organization, and mask selection.

\section{Conclusion}
\label{sec:Conclusion}
This work makes the role of MLLMs in SOD measurable and actionable. SaliLLM reveals that foundation models already excel at salient entity selection and coarse localization, while the remaining gap stems mainly from mismatches in foreground cardinality, granularity, and extent. We therefore reformulate zero-shot SOD as protocol conditioned foreground organization and introduce FOCUS, which combines top-down observer protocol calibration with bottom-up perceptual propagation to produce coherent prompts for a general segmenter. Without SOD specific training, FOCUS achieves the strongest overall or average performance among the compared methods across 13 datasets. These results establish foreground organization as the key link between MLLM perception and pixel-level saliency prediction, shifting SOD from learning dataset specific masks to organizing general visual knowledge under observer protocols.

\bibliography{aaai2027}


\end{document}